\documentclass[lettersize,journal]{IEEEtran}
\usepackage{amsmath,amsfonts}
\usepackage{amssymb,amsthm}
\usepackage{algorithmic}
\usepackage{algorithm}
\usepackage{array}
\usepackage{siunitx}
\usepackage{textcomp}
\usepackage{stfloats}
\usepackage{url}
\usepackage{verbatim}
\usepackage{graphicx}
\usepackage{makecell}
\usepackage{subfigure}
\usepackage{mathrsfs}
\usepackage[numbers]{natbib}
\usepackage{natbib}
\usepackage{booktabs}
\usepackage{hyperref}

\newcommand{\mytheoremname}{\bfseries Theorem}
\newcommand{\mylemmaname}{\bfseries Lemma}
\newcommand{\mydefinitionname}{\bfseries Definition}

\newtheorem{theorem}{\mytheoremname}
\newtheorem{lemma}[theorem]{\mylemmaname} % 共享编号
\newtheorem{definition}[theorem]{\mydefinitionname} % 共享编号

\begin{document}

\title{FlexiDrop: Theoretical Insights and Practical Advances in Random Dropout Method on GNNs}

\author{Zhiheng Zhou, Sihao Liu, Weichen Zhao\thanks{Zhiheng Zhou and Sihao Liu are the co-first authors. (Corresponding authors: Weichen Zhao.)

Zhiheng Zhou is with the Academy of Mathematics and Systems Science, Chinese Academy of Sciences and School of Mathematical Sciences, University of Chinese Academy of Sciences, Beijing, China. Email: zhouzhiheng@amss.ac.cn

Sihao Liu is with the School of Mathematics and Statistics, Guangdong University of Technology, Guangzhou, China. Email: amihua@mail2.gdut.edu.cn

Weichen Zhao is with the School of Statistics and Data Science, LPMC \& KLMDASR, Nankai University, Tianjin, China. Email: zhaoweichen@nankai.edu.cn}}% <-this % stops a space

% \thanks{Manuscript received April 19, 2021; revised August 16, 2021.}

% The paper headers
% \markboth{Journal of \LaTeX\ Class Files,~Vol.~14, No.~8, August~2021}%
% {Shell \MakeLowercase{\textit{et al.}}: A Sample Article Using IEEEtran.cls for IEEE Journals}

%\IEEEpubid{0000--0000/00\$00.00~\copyright~2021 IEEE}
% Remember, if you use this you must call \IEEEpubidadjcol in the second
% column for its text to clear the IEEEpubid mark.

\maketitle

\begin{abstract}
Graph Neural Networks (GNNs) are powerful tools for handling graph-type data. Recently, GNNs have been widely applied in various domains, but they also face some issues, such as overfitting, over-smoothing and non-robustness. The existing research indicates that random dropout methods are an effective way to address these issues. However, random dropout methods in GNNs still face unresolved problems. Currently, the choice of dropout rate, often determined by heuristic or grid search methods, can increase the generalization error, contradicting the principal aims of dropout. In this paper, we propose a novel random dropout method for GNNs called FlexiDrop. First, we conduct a theoretical analysis of dropout in GNNs using rademacher complexity and demonstrate that the generalization error of traditional random dropout methods is constrained by a function related to the dropout rate. Subsequently, we use this function as a regularizer to unify the dropout rate and empirical loss within a single loss function, optimizing them simultaneously. Therefore, our method enables adaptive adjustment of the dropout rate and theoretically balances the trade-off between model complexity and generalization ability. Furthermore, extensive experimental results on benchmark datasets show that FlexiDrop outperforms traditional random dropout methods in GNNs.
\end{abstract}


\section{Introduction}
Graph neural networks (GNNs), as deep learning models for processing structured data, have demonstrated successful applications in various scientific domains including biomedical~\cite{rhee2017hybrid,zitnik2018modeling,wang2020toward}, protein science~\cite{fout2017protein,strokach2020fast}, and combinatorial optimization~\cite{schuetz2022combinatorial,cappart2023combinatorial}. Moreover, they have played a significant role in multiple data mining scenarios such as recommendation systems~\cite{ying2018graph,fan2019graph,zhou2022decoupled}, traffic prediction~\cite{chen2019gated,ma2020streaming,li2021spatial}, and natural language processing~\cite{peng2018large,yao2019graph,liu2020tensor}.

However, GNNs may cause over-fitting, reducing their generalization ability~\cite{xue2021cap}. Furthermore, node representations tend to become indistinguishable due to the message-passing mechanism aggregating information from neighbors which called \emph{over-smoothing}. This phenomenon prevents the construction of deep GNNs~\cite{chen2020measuring}. Graph neural networks based on recursive structures are highly sensitive to the input graph data, and noise on the graph as well as malicious attacks can significantly impact the performance of GNNs~\cite{zhu2019robust}.

To enhance the generalization ability of GNNs, mitigate the over-smoothing phenomenon and increase the robustness of GNNs, researchers have proposed a random dropout method ~\cite{feng2020graph} for training GNN models. This method augmentes input data by randomly dropping some nodes, due to its simplicity and effectiveness, has been widely adopted in deep learning.

\begin{figure*}[h]
	\centering
	\subfigure[Cora] {
		\centering
		\includegraphics[width=0.95\columnwidth]{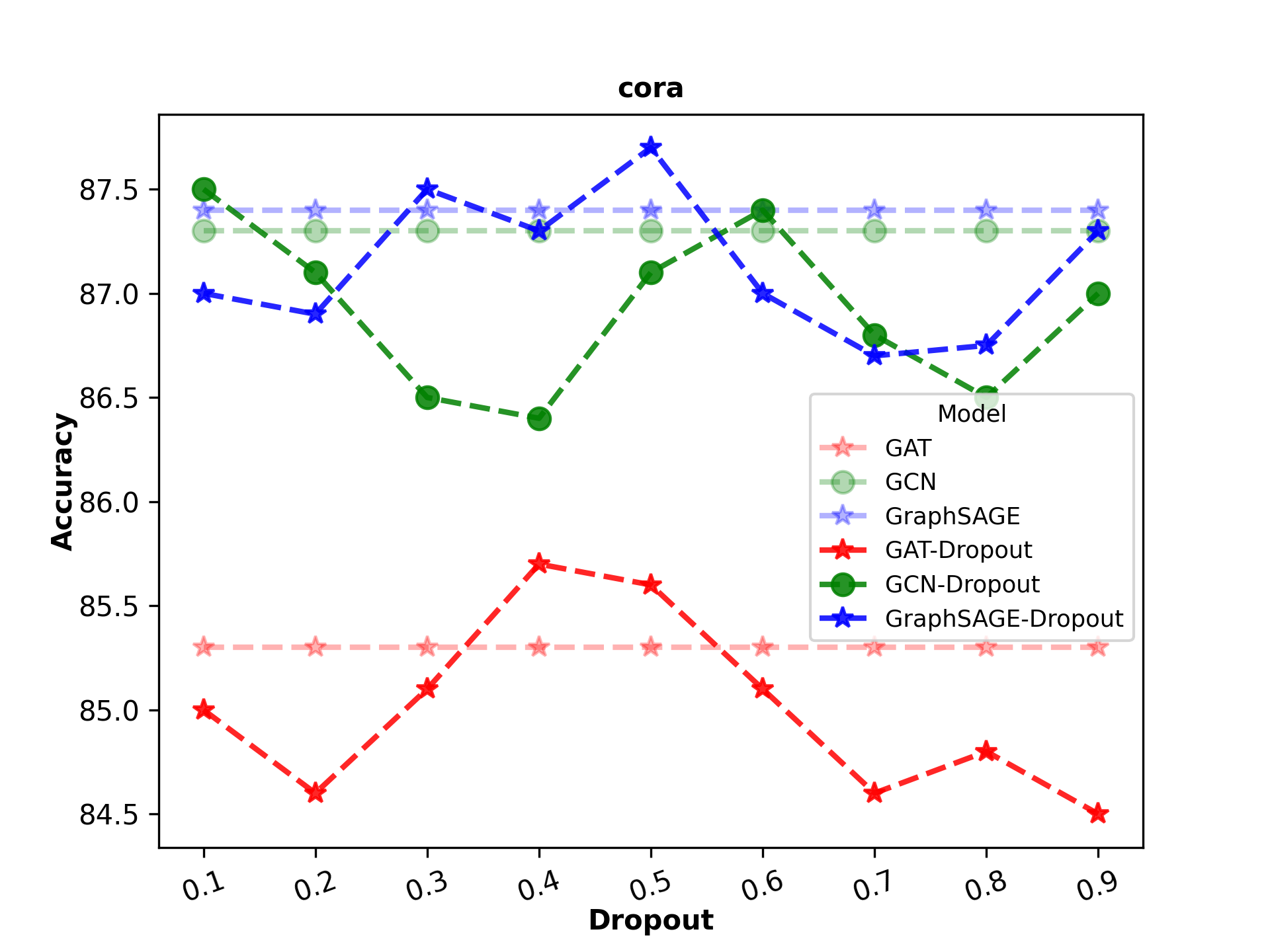}
        }
	\subfigure[PubMed] {
		\centering
		\includegraphics[width=0.95\columnwidth]{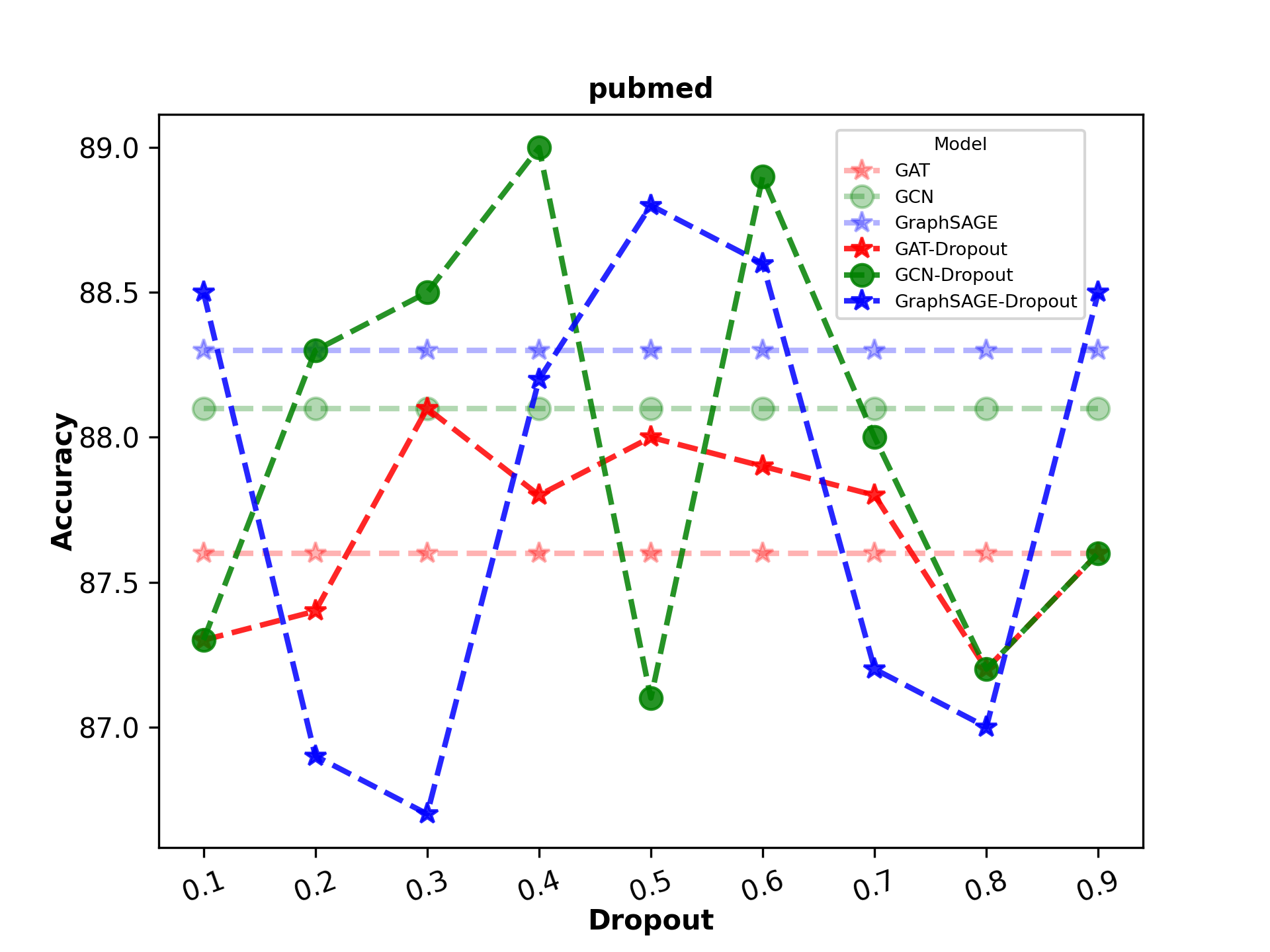}
		}
	\caption{The experimental results of three GNN backbone models and their dropout-based methods on the Cora and PubMed dataset with different dropout rates indicate that the optimal dropout rates differ among the models. Moreover, the performance of model at some dropout rates is inferior to that of the backbone models.}\label{Introduction}
\end{figure*}

Although the random dropout method has shown success in many GNNs application scenarios, a fundamental issue with existing methods is their lack of theoretical analysis. The answer to how to choose a proper dropout rate when applying these methods is still unclear. A fixed dropout rate is necessary during the training of models. However, determining the optimal dropout rate often involves a grid search across the dropout rate parameter space. As shown in Figure \ref{Introduction}, the optimal dropout rate varies across different models for the same datasets, and some inappropriate dropout rates can increase the generalization error, leading to a decrease in model performance. Moreover, within a given GNN layer, all inputs share the same dropout rate, despite different features contributing variably to the prediction.

To address the aforementioned limitations, we propose a novel random dropout method for GNNs in this paper, called FlexiDrop. It achieves adaptive adjustment of dropout rates and applies different rates to different features. We first prove that the generalization error of GNNs trained with dropout is constrained by a function related to the dropout rate. Subsequently, our theoretical derivation shows that optimizing generalization error equates to minimizing this constraint function. By integrating the constraint and empirical loss functions into a single loss function, we streamline the optimization process. This approach allows us to simultaneously target both generalization error and empirical loss, enhancing the model's performance. Subsequently, we introduce a method for adaptively adjusting the dropout rate in GNNs and provide theoretical guarantees that the method can improve the performance of GNNs. Finally, we conducted extensive experiments on benchmark graph datasets and compared FlexiDrop with baseline and dropout-based GNN models. Our results reveal that FlexiDrop enhances prediction accuracy, reduces over-smoothing, and improves robustness. These results confirm the effectiveness of our approach.

The main contributions of this paper are as follows:
\begin{itemize}
    \item We first prove that the generalization error in GNNs trained with dropout is constrained by a function related to the dropout rate. We achieve the goal of optimizing the model towards both generalization error and empirical loss by integrating the constraint function and the empirical loss function into a single loss function and optimizing them simultaneously.
    \item We introduce FlexiDrop as a novel GNN random dropout method that achieves adaptive adjustment of the dropout rate. Moreover, we provide theoretical guarantees that FlexiDrop can improve the performance of GNNs.
    \item We conduct experiments on various graph tasks using graph benchmark datasets. The experimental results show that FlexiDrop can better improve the performance of GNNs, alleviate over-smoothing and enhance robustness than both the backbone models and GNN models with dropout, which proves the effectiveness of our method.
\end{itemize}

\section{Related Work}
The trend of using non-Euclidean graphs to represent data has significantly increased in recent years. GNNs have become key to deploying deep learning on these graphs, achieving remarkable success across various fields~\cite{zhou2020graph}. GNNs adeptly capture intricate interconnections by aggregating and propagating messages among nodes. As graphs are increasingly used in modeling real-world scenarios, GNNs have shown their versatility in numerous applications.

\paragraph{\textbf{Limitations of GNNs}} Although GNNs excel in domains like node classification~\cite{sun2020adagcn}, graph classification~\cite{zhang2018end}, and link prediction~\cite{zhou2022decoupled}, they face challenges with complex graph data. A primary issue is overfitting, where the network performs well on training data but poorly on unseen data~\cite{hu2019strategies}. GNNs also encounter the over-smoothing challenge. This occurs when updates cause node features to become too similar, mirroring the effect seen in multi-layer GCNs that aggregate neighbor information. Consequently, nodes lose their distinctiveness~\cite{chen2020measuring}. Lastly, GNNs are limited in robustness, being sensitive to noisy data and adversarial attacks~\cite{gunnemann2022graph}. Therefore, despite their strong performance in many applications, these challenges limit their practicality and effectiveness.

\paragraph{\textbf{Dropout}} Dropout is now widely used in deep learning to overcome challenges related to over-fitting. Hinton et al.~\cite{hinton2012improving} first used Dropout as an effective heuristic for regularisation in CNNs, achieving higher accuracy than the original model on test data. 
Dropout and its varieties have achieved significant success across various model types, including CNNs~\cite{park2017analysis, singh2022dropout}, RNNs~\cite{gal2016theoretically, billa2018dropout}, Transformers~\cite{fan2019reducing}, and others~\cite{kim2019message,mordido2018dropout}. 
Essentially, random dropout is a strategy to introduce noise to features, aimed at reducing over-fitting by purposefully manipulating the training data~\cite{srivastava2014dropout}.

\paragraph{\textbf{Dropout with GNNs}} Despite GNNs and CNNs having similar structures, such as convolution and fully connected layers, their input formats and convolution processes differ significantly~\cite{daigavane2021understanding}. Research~\cite{shu2022understanding} suggests that dropout can also improve generalization in GNNs. Nevertheless, the most striking difference between GNNs and other neural networks like CNNs and RNNs from the nature of input data~\cite{zhou2020graph}. This implies that dropout strategies used for CNNs ought not to be straightforwardly transferred to GNNs without understanding their effects.

Papp et al.~\cite{papp2021dropgnn} introduced DropGNNs, a method employing selective node dropouts across multiple runs. This approach significantly enhances GNNs' representation abilities. It improves their capacity to distinguish complex graph structures, including neighborhoods that are indiscernible by traditional message-passing GNNs. Feng et al.~\cite{feng2020graph} introduced the GRAND model with a DropNode mechanism, utilizing a random propagation strategy and consistency regularization to enhance graph data augmentation, effectively mitigating over-smoothing and non-robustness, and demonstrating superior generalization compared to existing GNNs. Rong et al.~\cite{rong2019dropedge} presents DropEdge, a method enhancing GNNs by randomly removing edges during training. This technique improves both model performance and adaptability, validated through comprehensive tests. DropMessage~\cite{fang2023dropmessage} introduces a theoretical framework that enhances GNNs training stability and preserves feature diversity. This approach has been validated as effective across various datasets and models. Shu et al.~\cite{shu2022understanding} differentiated dropout strategies into two categories: feature map and graph structure dropouts. They proposed novel methods, including layer compensation and adaptive Gaussian techniques, to improve GNNs performance, especially in mitigating over-smoothing and enhancing transductive learning in deeper networks. These are shown to outperform baselines in shallow layers' GNNs, with varying effectiveness in transductive performance and over-smoothing mitigation, especially as GNNs become deeper.

A key difference between GNNs and traditional deep learning models is the over-smoothing of features as layer count increases~\cite{chen2020measuring}. 
Additionally, some work~\cite{fang2023dropmessage, rong2019dropedge, shu2022understanding} showed that dropout techniques notably reduce over-smoothing in GNNs. Identifying the optimal dropout rate is challenging~\cite{ba2013adaptive, zhai2018adaptive}. The process of adjusting dropout settings in GNNs is often manual and relies heavily on experience.

\section{Theoretical Analysis}
	In this section, we first introduced Rademacher complexity. Then, for GCNs, we demonstrated that the dropout method reduces the upper bound of generalization error based on Rademacher complexity.
	\subsection{Rademacher complexity}
	In learning theory, Rademacher complexity~\cite{mohri2018foundations,shalev2014understanding} is regarded as an important tool for studying the upper bound of generalization error of models. 
	\begin{definition}[Rademacher complexity]
		Let $\mathcal{H}$ be a set of hypothesis functions, $\mathcal{S}=\{\mathbf{x}_{1},\mathbf{x}_{2},\ldots,\mathbf{x}_{n}\}$ be a sample set, where $\mathbf{x}_{i}\sim D$ and $D$ is the distribution over the dataset $\mathcal{X}$. The \emph{Rademacher complexity} is defined as follows:
		\begin{equation}
			\mathfrak{R}_{\mathcal{S}}(\mathcal{H}):=\mathbb{E}\left[\sup_{h\in\mathcal{H}}\frac{1}{n}\sum_{i=1}^{n}\epsilon_{i}h(\mathbf{x}_{i})\right],
		\end{equation}
		where $\epsilon_{i}$ are independent random variables uniformly drawn from $\{1,-1\}$, which are referred to as \emph{Rademacher variables}. In a more general sense, for a set of vectors $\mathcal{A}\subset\mathbb{R}^{n}$, its Rademacher complexity is defined as follows:
		$$ \mathfrak{R}(\mathcal{A}):=\mathbb{E}\left[\sup_{\mathbf{a}\in \mathcal{A}}\frac{1}{n}\sum_{i=1}^{n}\epsilon_{i}a_{i}\right]. $$
	\end{definition}

	Rademacher complexity characterizes the richness of a hypothesis function family by measuring its adaptability to random noise, as it quantifies the extent to which a hypothesis set fits the Rademacher variables. The following Lemma \ref{lemma3.2} provides a general upper bound on the generalization error of learning problems based on Rademacher complexity. 
	\begin{lemma}\label{lemma3.2}[Theorem 3.1 of \cite{mohri2018foundations}]
		Let $\mathcal{F}=\{f(\mathbf{x},\mathbf{w}), \mathbf{w}\in\mathcal{W}\}$ be a set of hypothesis functions, $\ell$ be a loss function bounded by $B$, and $\mathcal{S}=\{\mathbf{x_1},\ldots,\mathbf{x_n}\}$ be a training sample, where $\mathbf{x}_{i}\sim D$ and $D$ is the distribution over $\mathcal{X}$. For any $\delta > 0$ and $\mathbf{w}\in\mathcal{W}$, the following inequality holds with probability at least $1-\delta$:
		\begin{equation}
		\begin{split}
		\mathbb{E}_{\mathbf{x}\sim D}[\ell(f(\mathbf{x},\mathbf{w}),y)]\leq &\frac{1}{n}\sum_{i=1}^{n}\ell(f(\mathbf{x}_{i},\mathbf{w}),y_{i}) \\
		& +2\mathfrak{R}_{\mathcal{S}}(\ell\circ\mathcal{F})+3B\sqrt{\frac{\ln \frac{2}{\delta}}{n}}.
		\end{split}
		\label{eq_rade}
		\end{equation}
	\end{lemma}
	\subsection{Dropout in graph neural networks}
	\textbf{Setup.} In this subsection, we consider the single-layer graph convolution networks without activation function as an example. By applying the same approach, the conclusions can be naturally extended to other graph neural networks. We provide a study in more general setup in section \ref{sec.4}. Let
	$$ \mathcal{S}=\left\{(\mathbf{x}_{u},y_{u})\;|\;\mathbf{x}_{u}\in\mathbb{R}^{d},y_{u}\in\{0,1\},u\in\mathcal{V}\right\} $$
	be input dataset, where $\mathbf{x}_u$ represents the features of node $u$, and $y_u$ represents the one-hot label of node $u$. Let $\ell$ be the loss function. The single-layer GCN can be represented as
	$$ F=PX\mathbf{w}, $$
	where $P := \tilde{D}^{-\frac{1}{2}}\tilde{A}\tilde{D}^{-\frac{1}{2}}=(P_{u,v})_{u,v\in\mathcal{V}}$ is the graph convolution operator, $X\in\mathbb{R}^{N\times d}$ is the feature matrix consisting of node feature vectors, and $\mathbf{w}=(w_{1},w_{2},\ldots,w_{d})\in\mathbb{R}^{d}$ is the weight vector. In component form, for node $u$, the output of the network is
	$$f(\mathbf{x}_{u};\mathbf{w})=\sum_{k=1}^{d}\left(\sum_{v=1}^{N}P_{u,v}x_{v,k}\right)w_{k}.$$
	The output of node $u$ with the dropout methed is
	$$f(\mathbf{x}_{u};\mathbf{w};p)=\mathbb{E}_{\Theta}\left[\sum_{k=1}^{d}\left(\sum_{v=1}^{N}P_{u,v}(\theta_{v,k}\cdot x_{v,k})\right)w_{k}\right],$$
	where $ \Theta:=(\theta_{i,j})\in\mathbb{R}^{N\times d} $, $\theta_{i,j}\sim \text{Bernoulli}(p)$ represents a Bernoulli random variable with parameter $ p $. The hypothesis function set is
	$$ \mathcal{F}_{\text{Dropout}}=\{f(\mathbf{x}_{u};\mathbf{w};p),\mathbf{w}\in\mathcal{W}\}, $$
	where $ \mathcal{W} $ is the parameter space, and the set of loss function is
	$$ \ell\circ\mathcal{F}_{\text{Dropout}}=\{((\mathbf{x}_{u},y_{u});p)\rightarrow \ell(f(\mathbf{x}_{u};\mathbf{w};p),y_{u}),\mathbf{w}\in\mathcal{W}\}, $$
	where $ \ell $ is loss function.
	
	\textbf{Upper bound of generalization error.} Based on the setup in this section, Theorem \ref{thm3.3} below provides an upper bound on the Rademacher complexity of the hypothesis function set for the dropout method.
	\begin{theorem}\label{thm3.3}
		Let $\mathcal{F}=\{f(\mathbf{x}_{u};\mathbf{w}),\mathbf{w}\in\mathcal{W}\}$, where the vector norm of $\mathbf{w}$ is bounded by a constant $B_{1}$, i.e., $\mathcal{W}:=\{\mathbf{w},\|\mathbf{w}\|_{2}\leq B_{1}\}$. The vector norm of each input $\mathbf{x}_{u}$ at each node is bounded by a constant $B_{2}$, and each row vector of $P$ is bounded by a constant $B_{3}$. Then,
		\begin{equation}\label{eq_thm3}
			\mathfrak{R}_{\mathcal{S}}(\mathcal{F}_{\text{Dropout}})\leq \frac{p}{\sqrt{N}}B_{1}B_{2}B_{3},\\
			\quad\mathfrak{R}_{\mathcal{S}}(\mathcal{F})\leq \frac{1}{\sqrt{N}}B_{1}B_{2}B_{3}.
		\end{equation}
		
	\end{theorem}

 \begin{proof}
	From the definition,
	$$ \begin{aligned}
	\mathfrak{R}_{\mathcal{S}}(\mathcal{F})&=\frac{1}{N}\mathbb{E}_{\epsilon}[\sup_{\mathbf{w}}\sum_{u=1}^{N}\epsilon_{u}f(\mathbf{x}_{u};\mathbf{w})]\\
	&=\frac{1}{N}\mathbb{E}_{\epsilon}[\sup_{\mathbf{w}}\sum_{u=1}^{N}\epsilon_{u}\sum_{k=1}^{d}(\sum_{v=1}^{N}P_{u,v}x_{v,k})w_{k}]\\
	&=\frac{1}{N}\mathbb{E}_{\epsilon}[\sup_{\mathbf{w}}\sum_{k=1}^{d}(\sum_{u=1}^{N}\epsilon_{u}\sum_{v=1}^{N}P_{u,v}x_{v,k})w_{k}].\\
	\end{aligned} $$
	By Cauchy-Schwarz inequality,
	$$ \begin{aligned}
	\frac{1}{N}\mathbb{E}_{\epsilon}[\sup_{\mathbf{w}}\sum_{k=1}^{d}&(\sum_{u=1}^{N}\epsilon_{u}\sum_{v=1}^{N}P_{u,v}x_{v,k})w_{k}]\\
	\leq\frac{\|\mathbf{w}\|}{N}\mathbb{E}_{\epsilon}[(\sum_{u=1}^{N}&\sum_{u'=1}^{N}\epsilon_{u}\epsilon_{u'}\\&(\sum_{k=1}^{d}(\sum_{v=1}^{N}P_{u,v}x_{v,k})\cdot(\sum_{v=1}^{N}P_{u',v}x_{v,k})))^{\frac{1}{2}}].\\
	\end{aligned} $$
	By Jensen inequality,
	$$ \begin{aligned}
	\leq\frac{\|\mathbf{w}\|}{N}(\sum_{u=1}^{N}&\sum_{u'=1}^{N}\mathbb{E}_{\epsilon}\epsilon_{u}\epsilon_{u'}\\&(\sum_{k=1}^{d}(\sum_{v=1}^{N}P_{u,v}x_{v,k})\cdot(\sum_{v=1}^{N}P_{u',v}x_{v,k})))^{\frac{1}{2}}.
	\end{aligned} $$
	Since for $ \epsilon_{i} $, if $ u=u' $, $ \mathbb{E}_{\epsilon}\epsilon_{u}\epsilon_{u'}=1 $, otherwise $ \mathbb{E}_{\epsilon}\epsilon_{u}\epsilon_{u'}=0 $,
	$$ \begin{aligned}
	\frac{\|\mathbf{w}\|}{N}(\sum_{u=1}^{N}&\sum_{u'=1}^{N}\mathbb{E}_{\epsilon}\epsilon_{u}\epsilon_{u'}\\&(\sum_{k=1}^{d}(\sum_{v=1}^{N}P_{u,v}x_{v,k})\cdot(\sum_{v=1}^{N}P_{u',v}x_{v,k})))^{\frac{1}{2}}\\=\frac{\|\mathbf{w}\|}{N}(&\sum_{u=1}^{N}\sum_{k=1}^{d}\sum_{v=1}^{N}(P_{u,v}x_{v,k})^{2})^{\frac{1}{2}}\\\leq\frac{1}{\sqrt{N}}&B_{1}B_{2}B_{3}.
	\end{aligned}  $$
	Therefore,
	$$\mathfrak{R}_{\mathcal{S}}(\mathcal{F})\leq \frac{1}{\sqrt{N}}B_{1}B_{2}B_{3}. $$
	Similarly, it can be inferred that
	$$ \mathfrak{R}_{\mathcal{S}}(\mathcal{F}_{\text{Dropout}})\leq \frac{p}{\sqrt{N}}B_{1}B_{2}B_{3}. $$
\end{proof}
 In classification, regression tasks and etc., loss functions such as cross-entropy loss are Lipschitz continuous. Regarding the Rademacher complexity of Lipschitz functions, the following lemma holds.
	\begin{lemma}\label{lemma2.13}[Contraction Lemma \cite{shalev2014understanding}]
		For each $i\in{1,2,\ldots,n}$, let $\phi_{i}:\mathbb{R}\rightarrow\mathbb{R}$ be a $\rho$-Lipschitz function. For $\mathbf{a}\in\mathbb{R}^{n}$, let $\mathbf{\phi}(\mathbf{a})$ denote the vector $(\phi_{1}(a_{1}),\phi_{2}(a_{2}),\ldots,\phi_{n}(a_{n}))$. Define $\mathbf{\phi}\circ \mathcal{A}:={\mathbf{\phi}(\mathbf{a}):\mathbf{a}\in \mathcal{A}}$. Then,
		\begin{equation}
			\mathfrak{R}(\mathbf{\phi}\circ \mathcal{A})\leq\rho\; \mathfrak{R}(\mathcal{A})
		\end{equation}
	\end{lemma}
	Let $\ell$ is $\rho$-Lipschitz, then by lemma \ref{lemma2.13},
	$$\mathfrak{R}_{\mathcal{S}}( \ell\circ\mathcal{F}_{\text{Dropout}})\leq\rho\; \mathfrak{R}_{\mathcal{S}}(\mathcal{F}_{\text{Dropout}}), $$
	$$\mathfrak{R}_{\mathcal{S}}( \ell\circ\mathcal{F})\leq\rho\; \mathfrak{R}_{\mathcal{S}}(\mathcal{F}). $$
	Therefore, the Rademacher complexity term in \eqref{eq_rade} can be written in a form independent of the loss function, that is, it can be expressed using $\mathfrak{R}_{\mathcal{S}}(\mathcal{F}_{\text{Dropout}})$ and $\mathfrak{R}_{\mathcal{S}}(\mathcal{F})$ as upper bounds on the generalization error. By combining this result with the conclusion from Theorem \ref{thm3.3}, noting that $p<1$ in \eqref{eq_thm3}, we can conclude that the dropout method can reduce the generalization upper bound of graph convolutional networks based on Rademacher complexity. Notice that our conclusion only concerns the boundedness condition, the conclusion still holds for other message passing neural networks such as graph attention networks.
	
	\section{FlexiDrop}\label{sec.4}
\begin{figure*}
    \centering
    \includegraphics[width=16cm]{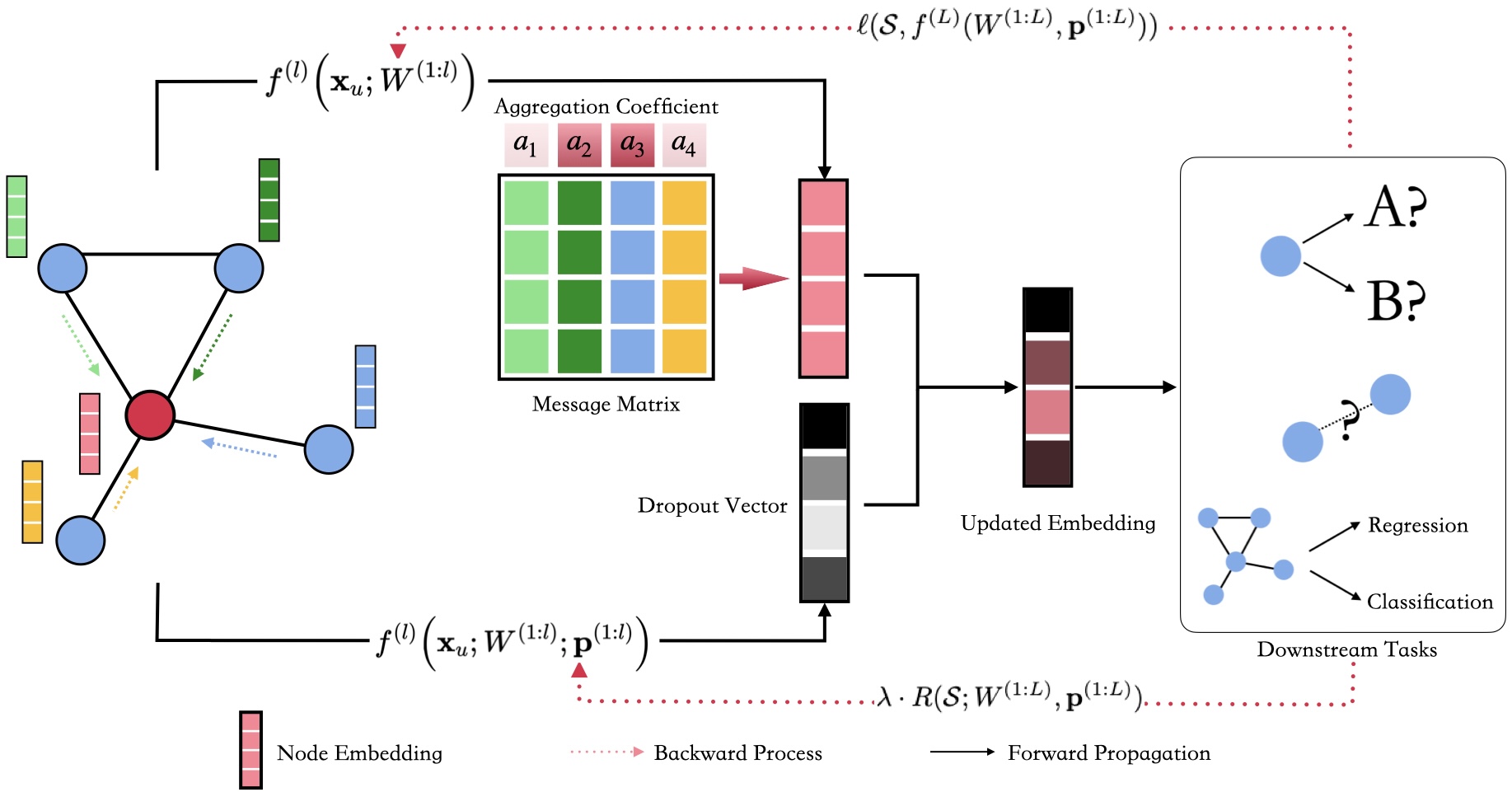}
    \caption{Framework of FlexiDrop. Initially, node embeddings generate the initial features and message passing occurs across the graph, producing a message matrix. Then, messages are weighted and aggregated using aggregation coefficients to form center node embeddings. Simultaneously, the model generates an adaptive dropout vector representing the retention probability of node embeddings at each layer. Finally, these updated embeddings are used for downstream tasks. Throughout the process, dropout rates are adaptively learned through forward and backward propagation to optimize task performance.}
    \label{Model}
\end{figure*}
	In this section, we extend the generalization bounds studied in the previous section to a more general case, investigating the generalization error bounds based on Rademacher complexity in general form GNNs with dropout. Furthermore, based on the derived generalization error bounds, we propose FlexiDrop, a random dropout method for GNNs that enables the flexible adaptation of dropout rates.
	\subsection{Setup}
	Let $\mathcal{S}=\left\{(\mathbf{x}_{u},\mathbf{y}_{u})\;|\;\mathbf{x}_{u}\in\mathbb{R}^{d},\mathbf{y}_{u}\in\{0,1\}^{C},u\in\mathcal{V}\right\} $, where $\mathbf{x}_{u}$ represents the features of node $u$ and $\mathbf{y}_{u}$ represents the one-hot class label of node $u$. $C$ denotes the number of predicted classes. Considering a general $L$-layer graph neural network model, let $ W^{(l)}\in\mathbb{R}^{k^{(l-1)}\times k^{(l)}} $ be the weight matrix from the $(l-1)$-th layer to the $l$-th layer, where $ k^{(0)}=d $ and $k^{(L)}=C$. The parameter set is denoted as $W^{(1:l)}:=\{W^{(1)},\ldots,W^{(l)}\}$. $W^{(l)}(:,j),j=1,\ldots,k^{(l)}$ represents the vector composed of the $j$-th column of $W^{(l)}$. The $j$-th component of the output of node $u$ before being activated in the $l$-th layer is represented as
	\begin{align*}
	f_{j}^{(l)}&(\mathbf{x}_{u};W^{(1:l)}):=\\&\sum_{k=1}^{k^{(l-1)}}\left(\sum_{v=1}^{N}a^{(l)}_{u,v}\sigma\left(f_{k}^{(l-1)}(\mathbf{x}_{n};W^{(1:l-1)})\right)\right)w^{(l)}_{k,j},
	\end{align*}
	where $a^{(l)}_{u,v}\ge 0$ represents the aggregation coefficient of the $l$-th layer, and without loss of generality, for $ \sum_{v=1}^{N}a^{(l)}_{u,v}=1 $, $ \sigma $ denotes the activation function, such as ReLU or others. Considering a classification problem, we restate the cross-entropy loss based on the Softmax function as follows
	$$ \ell(f^{(L)}(\mathbf{x}_{u};W^{(1:L)}),\mathbf{y})=-\sum_{j=1}^{k^{(L)}}y_{j}\log\frac{e^{f_{j}^{(L)}}}{\sum_{j}e^{f_{j}^{(L)}}}. $$
	
	Let $\mathbf{r}^{(l)}\in\{0,1\}^{k^{(l)}}$ be a random vector generated by $k^{(l)}$ independent Bernoulli random variables, that is, $r_{j}^{(l)}\sim\text{Bernoulli}(p_{j}^{(l)})$, where $1-p_{j}^{(l)}$ represents the dropout rate. We denote $ \mathbf{r}^{(1:l)}:=\{\mathbf{r}^{(1)},\ldots,\mathbf{r}^{(l)}\} $ and $ \mathbf{p}^{(1:l)}:=\{\mathbf{p}^{(1)},\ldots,\mathbf{p}^{(l)}\} $. The $j$-th component of the output of node $u$ before being activated in the $l$-th layer is expressed as
	\begin{align*}
	&f_{j}^{(l)}(\mathbf{x}_{u};W^{(1:l)};\mathbf{r}^{(1:l)}):=\\&\sum_{k=1}^{k^{(l-1)}}\left(\sum_{v=1}^{N}a^{(l)}_{u,v}r_{k}^{(l)}\sigma\left(f_{k}^{(l-1)}(\mathbf{x}_{v};W^{(1:l-1)};\mathbf{r}^{(1:l-1)})\right)\right)w^{(l)}_{k,j}.
	\end{align*}
	The output of node $ u $ is 
	\begin{equation*}
		f^{(L)}(\mathbf{x}_{u};W^{(1:L)};\mathbf{p}^{(1:L)}):=\mathbb{E}_{\mathbf{r}^{(1:L)}}\left[f^{(L)}(\mathbf{x}_{u};W^{(1:L)};\mathbf{r}^{(1:L)})\right].
	\end{equation*}
	\subsection{Upper bound of generalization error}
	The following Theorem \ref{thm4.1} provides an upper bound of the Rademacher complexity of the loss function for general $L$-layer GNNs using the dropout method.
	\begin{theorem}\label{thm4.1}
		Let $$ \mathcal{W}:=\left\{W^{(1:L)}|\max_{j}\|W^{(l)}(:,j)\|_{2}\leq B^{(l)},\; l=1,\ldots,L\right\}, $$ 
		and
		$$ \ell\circ f^{(L)}:=\left\{(\mathbf{x},\mathbf{y})\rightarrow\ell(f^{(L)}(\mathbf{x};W^{(1:L)};\mathbf{p}^{(1:L)}),\mathbf{y})\right\}, $$ then
		\begin{equation}\label{eq:rade0}
			\begin{split}
			\mathfrak{R}_{\mathcal{S}}(\ell\circ& f^{(L)})\leq M\cdot(\prod_{l=1}^{L}B^{(l)}\|\mathbf{p}^{(l)}\|_{2}),
			\end{split}
		\end{equation}
		where $M:=2^{L}C(\frac{2\log(2d)}{N})^{\frac{1}{2}}\cdot\max_{u}\|\mathbf{x}_{u}\|_{\infty}$ and for $\mathbf{x}\in\mathbb{R}^d$, $\|\mathbf{x}\|_{\infty}:=\max\{|x_1|,|x_2|,\ldots,|x_F|\}$.
	\end{theorem}

The proof of Theorem \ref{thm4.1} relies on the following lemmas from~\cite{shalev2014understanding}.
\begin{lemma}\label{lem2.12}
	Let $ \mathcal{A} $ be a subset of $ \mathbb{R}^{m} $,
	\begin{equation}
		\mathcal{A}':=\{\sum_{j=1}^{N}\alpha_{j}\mathbf{a}^{(j)}:N\in\mathbb{N},\;\mathbf{a}^{(j)}\in \mathcal{A},\;\alpha_{j}\ge 0,\;\sum_{j}\alpha_{j}=1\}.
	\end{equation}
	Then $ \mathfrak{R}(\mathcal{A}')=\mathfrak{R}(\mathcal{A}) $.
\end{lemma}
\begin{lemma}\label{lem2.13}
	For all $ i\in\{1,2,\ldots,n\} $, let $ \phi_{i}:\mathbb{R}\rightarrow\mathbb{R} $ be a $ \rho- $Lipschitz function. For $ \mathbf{a}\in\mathbb{R}^{n} $, denote $ \mathbf{\phi}(\mathbf{a}) := (\phi_{1}(a_{1}),\ldots,\phi_{n}(a_{n})) $. Let $ \mathbf{\phi}\circ \mathcal{A}:=\{\mathbf{\phi}(\mathbf{a}):\mathbf{a}\in \mathcal{A}\} $, then
	\begin{equation}
		\mathfrak{R}(\mathbf{\phi}\circ \mathcal{A})\leq\rho\; \mathfrak{R}(\mathcal{A}).
	\end{equation}
\end{lemma}
\begin{lemma}\label{lem2.14}
	Let $ \mathcal{S}=\{\mathbf{x}_{1},\ldots,\mathbf{x}_{n}\} $ be a subset of $ \mathbb{R}^{N} $, $ \mathcal{H}:=\{\mathbf{x}\mapsto\langle\mathbf{w},\mathbf{x}\rangle:\|\mathbf{w}\|_{1}\leq 1\} $, then
	\begin{equation}
		\mathfrak{R}_{\mathcal{S}}(\mathcal{H})\leq\max_{i}\|\mathbf{x}_{i}\|_{\infty}\left(\frac{2\log(2N)}{n}\right)^{\frac{1}{2}}.
	\end{equation}
\end{lemma}
\begin{proof}[Proof of Theorem \ref{thm4.1}]
	Let
	$$ \mathfrak{R}_{\mathcal{S}}(f^{(l)}):=\frac{1}{N}\mathbb{E}_{\epsilon}\sup_{W^{(1:l)}}\sum_{u=1}^{N}\epsilon_{u}\sup_{j}\mathbb{E}_{\mathbf{r}^{(1:l)}}\left[f_{j}^{(l)}(\mathbf{x}_{u};W^{(1:l)};\mathbf{r}^{(1:l)})\right]. $$
	From lemma \ref{lem2.13},
	\begin{equation}\label{eq:rade}
	\mathfrak{R}_{\mathcal{S}}(\ell\circ f^{(L)})\leq C\mathfrak{R}_{\mathcal{S}}(f^{(L)}).
	\end{equation}
	Let
	$$ \hat{\mathfrak{R}}_{\mathcal{S}}(f^{(l)}):=\mathbb{E}_{\mathbf{r}^{(1:l)}}[\frac{1}{N}\mathbb{E}_{\epsilon}\sup_{W^{(1:l)}}|\sum_{u=1}^{N}\epsilon_{u}\sup_{j}f_{j}^{(l)}(\mathbf{x}_{u};W^{(1:l)};\mathbf{r}^{(1:l)})|]. $$
	Then
	\begin{equation}\label{eq:rade1}
	\mathfrak{R}_{\mathcal{S}}(f^{(l)})\leq \hat{\mathfrak{R}}_{\mathcal{S}}(f^{(l)}).
	\end{equation}
	We have
	$$ \begin{aligned}
	&\hat{\mathfrak{R}}_{\mathcal{S}}(f^{(l)})\\
& =\mathbb{E}_{\mathbf{r}^{(1:l)}}[\frac{1}{N}\mathbb{E}_{\epsilon}\sup_{W^{(1:l)}}|\sum_{u=1}^{N}\epsilon_{u}\sup_{j}f_{j}^{(l)}(\mathbf{x}_{u};W^{(1:l)};\mathbf{r}^{(1:l)})|]\\
	&=\mathbb{E}_{\mathbf{r}^{(1:l)}}[\frac{1}{N}\mathbb{E}_{\epsilon}\sup_{W^{(1:l)}}|\sum_{u=1}^{N}\epsilon_{u}\\&\sup_{j}\sum_{k=1}^{k^{(l-1)}}(\sum_{v=1}^{N}a^{(l)}_{u,v}r_{k}^{(l)}\sigma(f_{k}^{(l-1)}(\mathbf{x}_{v};W^{(1:l-1)};\mathbf{r}^{(1:l-1)})))w^{(l)}_{k,j}|].
	\end{aligned} $$
	From lemma \ref{lem2.13},
	$$ \begin{aligned}
	\leq 2\mathbb{E}_{\mathbf{r}^{(1:l)}}[\frac{1}{N}\mathbb{E}_{\epsilon}&\sup_{W^{(1:l)}}|\sum_{u=1}^{N}\epsilon_{u}\\\sup_{j}\sum_{k=1}^{k^{(l-1)}}w_{k,j}^{(l)}&r_{k}^{(l)}(\sum_{v=1}^{N}a^{(l)}_{u,v}f_{k}^{(l-1)}(\mathbf{x}_{v};W^{(1:l-1)};\mathbf{r}^{(1:l-1)}))|]\\
	= 2\mathbb{E}_{\mathbf{r}^{(1:l)}}[\frac{1}{N}\mathbb{E}_{\epsilon}&\sup_{W^{(1:l)}}|\sup_{j}\\\sum_{k=1}^{k^{(l-1)}}w_{k,j}^{(l)}r_{k}^{(l)}&\sum_{u=1}^{N}\epsilon_{u}(\sum_{v=1}^{N}a^{(l)}_{u,v}f_{k}^{(l-1)}(\mathbf{x}_{v};W^{(1:l-1)};\mathbf{r}^{(1:l-1)}))|].
	\end{aligned} $$
	From Jensen inequality,
	\begin{align*}
	\leq 2\mathbb{E}_{\mathbf{r}^{(1:l)}}&[(\sup_{W^{(l)},j}\sum_{k=1}^{k^{(l-1)}}|w_{k,j}^{(l)}r_{k}^{(l)}|)\frac{1}{N}\mathbb{E}_{\epsilon}\sup_{W^{(1:l-1)}}\\&|\sum_{u=1}^{N}\epsilon_{u}(\sum_{v=1}^{N}a^{(l)}_{u,v}\sup_{j} f_{j}^{(l-1)}(\mathbf{x}_{v};W^{(1:l-1)};\mathbf{r}^{(1:l-1)}))|].
	\end{align*}
	From Cauchy-Schwarz inequality,
	\begin{align*}
		\leq 2\mathbb{E}_{\mathbf{r}^{(1:l)}}&[(\sup_{W^{(l)},j}\|W^{(l)}(:,j)\|_{2}\|\mathbf{r}^{(l)}\|_{2})\frac{1}{N}\mathbb{E}_{\epsilon}\sup_{W^{(:l-1)}}\\&|\sum_{u=1}^{N}\epsilon_{u}(\sum_{v=1}^{N}a^{(l)}_{u,v}\sup_{j} f_{j}^{(l-1)}(\mathbf{x}_{v};W^{(1:l-1)};\mathbf{r}^{(1:l-1)}))|]
	\end{align*}
	\begin{align*}
		\leq2B^{(l)}&\|\mathbf{p}^{(l)}\|_{2}\mathbb{E}_{\mathbf{r}^{(1:l-1)}}[\frac{1}{N}\mathbb{E}_{\epsilon}\sup_{W^{(1:l-1)}}\\&|\sum_{u=1}^{N}\epsilon_{u}(\sum_{nv=1}^{N}a^{(l)}_{u,v}\sup_{j} f_{j}^{(l-1)}(\mathbf{x}_{v};W^{(1:l-1)};\mathbf{r}^{(1:l-1)}))|].
	\end{align*}
	Since $\sum_{n=1}^{N}a^{(l)}_{u,v}=1$, from lemma \ref{lem2.12},
	\begin{equation}\label{eq:rade2}
	\hat{\mathfrak{R}}_{\mathcal{S}}(f^{(l)})\leq2B^{(l)}\|\mathbf{p}^{(l)}\|_{2}\hat{\mathfrak{R}}_{\mathcal{S}}(f^{(l-1)}).
	\end{equation}
	After obtaining the recursive relationship between the $l$-th layer and the $(l-1)$-th layer, we return to the first layer,
	$$ \begin{aligned}
	\hat{\mathfrak{R}}_{\mathcal{S}}(f^{(1)})&=\mathbb{E}_{\mathbf{r}^{(1)}}[\frac{1}{N}\mathbb{E}_{\epsilon}\sup_{W^{(1)}}|\sum_{u=1}^{N}\epsilon_{u}\sup_{j}f_{j}^{(1)}(\mathbf{x}_{u};W^{(1)};\mathbf{r}^{(1)})|]\\
	=\mathbb{E}_{\mathbf{r}^{(1)}}&[\frac{1}{N}\mathbb{E}_{\epsilon}\sup_{W^{(1)}}|\sum_{u=1}^{N}\epsilon_{u}\sup_{j}\sum_{k=1}^{d}w_{k,j}^{(1)}r_{k}^{(1)}(\sum_{v=1}^{N}a^{(1)}_{u,v}x_{v,k})|].\end{aligned} $$
	Let $ \odot $ denote Hadamard product, from Hölder's inequality,
	\begin{align*}
	&\leq \mathbb{E}_{\mathbf{r}^{(1)}}\left[\frac{1}{N}\mathbb{E}_{\epsilon}\sup_{W^{(1)},j}\|W^{(1)}(:,j)\odot\mathbf{r}^{(1)}\|_{1}\|\sum_{u=1}^{N}\epsilon_{u}\mathbf{x}_{u}\|_{\infty}\right]\\
	&\leq\frac{B^{(1)}\|\mathbf{p}^{(1)}\|_{2}}{N}\mathbb{E}_{\epsilon}\left[\|\sum_{u=1}^{N}\epsilon_{u}\mathbf{x}_{u}\|_{\infty}\right].
	\end{align*}
	From lemma \ref{lem2.14},
	\begin{equation}
		\frac{1}{N}\mathbb{E}_{\epsilon}\left[\|\sum_{u=1}^{N}\epsilon_{u}\mathbf{x}_{u}\|_{\infty}\right]\leq\max_{u}\|\mathbf{x}_{u}\|_{\infty}\left(\frac{2\log(2d)}{N}\right)^{\frac{1}{2}}.
	\end{equation}
	Then,
	\begin{equation}\label{eq:rade3}
	\hat{\mathfrak{R}}_{\mathcal{S}}(f^{(1)})\leq B^{(1)}\|\mathbf{p}^{(1)}\|_{2}\left(\frac{2\log(2d)}{N}\right)^{\frac{1}{2}}\max_{u}\|\mathbf{x}_{u}\|_{\infty}.
	\end{equation}
	By combining equations (\ref{eq:rade}), (\ref{eq:rade1}), (\ref{eq:rade2}), and (\ref{eq:rade3}), we can obtain
	\begin{align*}
	\mathfrak{R}_{\mathcal{S}}(\ell\circ& f^{(L)})\leq\\ &2^{L}C(\frac{2\log(2d)}{N})^{\frac{1}{2}}\max_{u}\|\mathbf{x}_{u}\|_{\infty}(\prod_{l=1}^{L}B^{(l)}\|\mathbf{p}^{(l)}\|_{2}).
	\end{align*}
\end{proof}	
 Combining Theorem \ref{thm4.1} with Lemma \ref{lemma3.2}, we can obtain the corresponding upper bound on the generalization error for general $L$-layer GNNs using the dropout.
	\subsection{FlexiDrop}
	Based on the findings of Theorem \ref{thm4.1}, we propose an adaptive random dropout method called FlexiDrop, which treats the dropout rate as a trainable parameter for adaptive optimization during training. This approach alleviates the challenges associated with hyperparameter tuning.
	
	Noting that the upper bound of the Rademacher complexity of the loss function in Equation (\ref{eq:rade0}) is related to the probabilities $\mathbf{p}^{(1:L)}$, in conjunction with Lemma \ref{lemma3.2}, the generalization error of a graph neural network model using the random dropout method is constrained by a function of $\mathbf{p}^{(1:L)}$. By utilizing a penalty function approach, we can treat $\mathbf{p}^{(1:L)}$ as a trainable parameter and incorporate it into the optimization objective. Through adaptive unconstrained optimization of $\mathbf{p}^{(1:L)}$, we can reduce the upper bound of the model's generalization error. Specifically, the complete objective function can be expressed as follows
	\begin{equation}
		\begin{split}
		\text{Obj}(W^{(1:L)},\mathbf{p}^{(1:L)}):=\ell&(\mathcal{S},f^{(L)}(W^{(1:L)},\mathbf{p}^{(1:L)}))\\
		& +\lambda\cdot R(\mathcal{S};W^{(1:L)},\mathbf{p}^{(1:L)}),
		\end{split}
	\end{equation}
	where 
	\begin{equation}
	\begin{split}
	R(&\mathcal{S};W^{(1:L)},\mathbf{p}^{(1:L)}):=M\cdot(\prod_{l=1}^{L}\max_{j}\|W^{(l)}(:,j)\|_{2}\|\mathbf{p}^{(l)}\|_{2}),
	\end{split}
	\end{equation}
	where $ M:=2^{L}C(\frac{2\log(2d)}{N})^{\frac{1}{2}}\cdot\max_{u}\|\mathbf{x}_{u}\|_{\infty} $. By optimizing the objective function $\text{Obj}(W^{(1:L)},\mathbf{p}^{(1:L)})$ using gradient descent algorithms, it is possible to reduce the Rademacher complexity \eqref{eq:rade0}. It improves the upper bound of the model's generalization error and allows for flexible adjustment of the dropout parameters.

\section{Experiment}
In this section, we first give the experimental setup, then evaluate the effectiveness of the FlexiDrop, compared with the backbone models and their dropout methods. Next, we give the parameter analysis. We finally give the over-smoothing analysis. 
We primarily explore the following questions: 
\begin{enumerate}
    \item Does FlexiDrop perform better than random dropout methods on GNNs?
    \item Does FlexiDrop address the issue of sensitivity to dropout rates in random dropout methods on GNNs?
    \item Does FlexiDrop mitigate the over-smoothing problem in GNNs?
    \item Can FlexiDrop effectively enhance the robustness of GNNs to resist attacks?
\end{enumerate}

\begin{table*}[htbp]
  \centering
  \caption{Summary of datasets.}
  \resizebox{18cm}{!}{ 
  \begin{tabular}{@{}lccccccr@{}}
    \toprule
    Dataset   & Graphs & Graph classes & Avg. nodes & Avg. edges & Node features & Node classes & Task (N/L/G) \\
    \midrule
    Cora      & 1      & -             & 2708       & 5429       & 1433          & 7            & N,L          \\
    Citeseer  & 1      & -             & 3312       & 4732       & 3703          & 6            & N,L          \\
    PubMed    & 1      & -             & 19717      & 44338      & 500           & 3            & N,L          \\
    ENZYMES   & 600    & 6             & 32.63      & 62.14      & 18            & -            & G            \\
    BZR       & 405    & 2             & 35.75      & 38.36      & 3             & -            & G            \\
    COX2      & 467    & 2             & 41.22      & 43.45      & 3             & -            & G            \\
    \bottomrule
  \end{tabular}
  }
  \label{table1}
\end{table*}

\begin{table}[b]
  \centering
  \caption{Comparison results of base models and dropout methods. The best results are in bold.}
  {
   \begin{tabular}{
      lccc
    }
    \toprule
    {model} & \multicolumn{3}{c}{Graph classification} \\
    \cmidrule(lr){2-4} & {BZR} & {COX2} & {ENZYMES} \\
    \midrule
    GCN          & 83.75$\pm$1.02 & 78.85$\pm$0.51 & 69.17$\pm$1.18 \\
    GCN-Dropout  & 86.03$\pm$0.55 & 80.65$\pm$0.78 & 75.87$\pm$1.62 \\
    GCN-Dropnode  & 84.68$\pm$0.45 & 79.99$\pm$1.02 & 73.14$\pm$2.01 \\
    GCN-Dropedge  & 87.33$\pm$0.63 & 80.82$\pm$0.77 & 75.88$\pm$1.65 \\
    GCN-FlexiDrop &  \textbf{89.17$\pm$0.56} & \textbf{82.44$\pm$0.51} & \textbf{77.22$\pm$1.08} \\
    \midrule
    GAT    & 85.42$\pm$1.18& 79.93$\pm$1.34 & 68.61$\pm$1.71 \\
    GAT-Dropout  & 87.42$\pm$1.41 & \textbf{81.90$\pm$1.65} & 73.34$\pm$1.71 \\
    GAT-Dropnode  & 87.55$\pm$0.98 & 80.89$\pm$1.53 & 75.33$\pm$1.92 \\
    GAT-Dropedge  & 86.31$\pm$1.22 & 81.24$\pm$1.64 & 74.68$\pm$2.23 \\
    GAT-FlexiDrop & \textbf{89.17$\pm$0.58} & 81.36$\pm$1.35 & \textbf{78.61$\pm$1.17} \\
    \midrule
    GraphSAGE & 85.13$\pm$0.62 & 80.04$\pm$0.47 & 70.13$\pm$1.55 \\
    GraphSAGE-Dropout  & 86.27$\pm$0.97 & 81.33$\pm$0.51 & 74.86$\pm$0.67 \\
    GraphSAGE-Dropnode  & 85.99$\pm$0.84 & 81.29$\pm$0.40 & 73.78$\pm$1.15 \\
    GraphSAGE-Dropedge  & 85.62$\pm$1.01 & 80.95$\pm$0.44 & 73.93$\pm$0.64 \\
    GraphSAGE-FlexiDrop & \textbf{88.96$\pm$1.28} & \textbf{82.80$\pm$0.88} & \textbf{78.00$\pm$0.71} \\
    \bottomrule
    \end{tabular}}
  \label{table_graph}
\end{table}

\begin{table*}[htbp]
  \centering
  \caption{Comparison results of base models and dropout methods. The best results are in bold.}
  \resizebox{\textwidth}{!}{
   \begin{tabular}{
      lcccccc
    }
    \toprule
    {model} & \multicolumn{3}{c}{Node classification} & \multicolumn{3}{c}{Link prediction}  \\
    \cmidrule(lr){2-4} \cmidrule(lr){5-7} 
    & {Cora} & {CiteSeer} & {PubMed} & {Cora} & {CiteSeer} & {PubMed}  \\
    \midrule
    GCN         &  87.30$\pm$0.37 & 78.47$\pm$0.62 & 88.10$\pm$0.51 & 80.54$\pm$0.08 & 80.14$\pm$0.06 & 78.93$\pm$0.08  \\
    GCN-Dropout  & 87.52$\pm$0.54 & 79.13$\pm$0.48 & 89.02$\pm$0.61 & 80.85$\pm$0.22 & 80.32$\pm$0.11 & 79.10$\pm$0.22  \\
    GCN-Dropnode  & 87.56$\pm$0.41 & 79.11$\pm$0.28 & 88.45$\pm$0.55 & 80.85$\pm$0.25 & 80.15$\pm$0.13 & 78.95$\pm$0.11  \\
    GCN-Dropedge  & 87.38$\pm$0.65 & 78.65$\pm$0.44 & 88.66$\pm$0.61 & 80.67$\pm$0.19 & 80.21$\pm$0.15 & 79.05$\pm$0.18 \\
    GCN-FlexiDrop & \textbf{87.97$\pm$0.24} & \textbf{79.57$\pm$0.31} & \textbf{89.30$\pm$0.38} & \textbf{80.99$\pm$0.15} & \textbf{80.40$\pm$0.05} & \textbf{79.59$\pm$0.14}  \\
    \midrule
    GAT    & 85.30$\pm$0.41 & 78.47$\pm$0.34 & 87.63$\pm$0.37 & 80.82$\pm$0.10 & 79.70$\pm$0.29 & 79.65$\pm$0.01  \\
    GAT-Dropout  & 85.74$\pm$0.79 & 78.86$\pm$0.73 & 88.07$\pm$0.85 & \textbf{81.07$\pm$0.19} & 79.88$\pm$0.16 & 79.72$\pm$0.09 \\
    GAT-Dropnode  & 85.96$\pm$0.62 & 79.12$\pm$0.50 & 87.96$\pm$0.44 & 80.93$\pm$0.12 & 80.12$\pm$0.17 & \textbf{79.81$\pm$0.11}  \\
    GAT-Dropedge  & 85.57$\pm$0.56 & 78.97$\pm$0.45 & 87.79$\pm$0.38 & 80.88$\pm$0.21 & 79.78$\pm$0.25 & 79.66$\pm$0.13 \\
    GAT-FlexiDrop& \textbf{87.27$\pm$0.47} & \textbf{79.35$\pm$0.33} & \textbf{88.34$\pm$0.42} & 81.00$\pm$0.16 & \textbf{80.68$\pm$0.13} & 79.78$\pm$0.07  \\
    \midrule
    GraphSAGE & 87.35$\pm$0.29 & 78.86$\pm$0.38 & 88.32$\pm$0.55  & 80.02$\pm$0.06 & 79.97$\pm$0.04 & 79.02$\pm$0.09 \\
    GraphSAGE-Dropout  & 87.74$\pm$0.34 & 79.03$\pm$0.62 & 88.79$\pm$0.43  & 80.41$\pm$0.11 & 80.15$\pm$0.14 & 79.28$\pm$0.05  \\
    GraphSAGE-Dropnode  & 87.58$\pm$0.28 & 79.11$\pm$0.45 & 89.03$\pm$0.61  & 80.33$\pm$0.09 & 80.08$\pm$0.10 & 79.37$\pm$0.08  \\
    GraphSAGE-Dropedge  & 87.41$\pm$0.69 & 78.99$\pm$0.69 & 89.15$\pm$0.36 & 80.10$\pm$0.16 & 80.01$\pm$0.14 & 79.16$\pm$0.18  \\
    GraphSAGE-FlexiDrop& \textbf{88.03$\pm$0.23} & \textbf{79.86$\pm$0.42} & \textbf{89.37$\pm$0.45} & \textbf{80.71$\pm$0.08} & \textbf{80.45$\pm$0.04} & \textbf{79.63$\pm$0.11}  \\
    \bottomrule
    \end{tabular}
  }
  \label{table2}
\end{table*}

\subsection{Experimental Setup}
\paragraph{\textbf{Datasets}} We employ six benchmark datasets for evaluation. Cora, CiteSeer and PubMed are citation networks~\cite{sen2008collective}. These three different citation networks are widely used as graph benchmarks. We conduct node classification and link prediction tasks on each dataset to determine the research area of papers/researchers and predict whether one paper cites another. ENZYMES~\cite{wang2022faith} is a dataset of 600 enzymes obtained from the BRENDA enzyme database. BZR~\cite{rossi2015network} is a dataset collected 405 ligands for benzodiazepine receptor. COX2~\cite{rossi2015network} is a dataset of molecular structures. We conduct graph tasks on each dataset.

We summarize these datasets in Table \ref{table1}. Note that the “Task” column indicates the
type of downstream task performed on each dataset: "N" for node classifcation, "L" for link prediction and "G" for graph classifcation.

\begin{figure*}[htbp]
\centering
	\subfigure[ACC] {
		\centering
		\includegraphics[width=0.85\columnwidth]{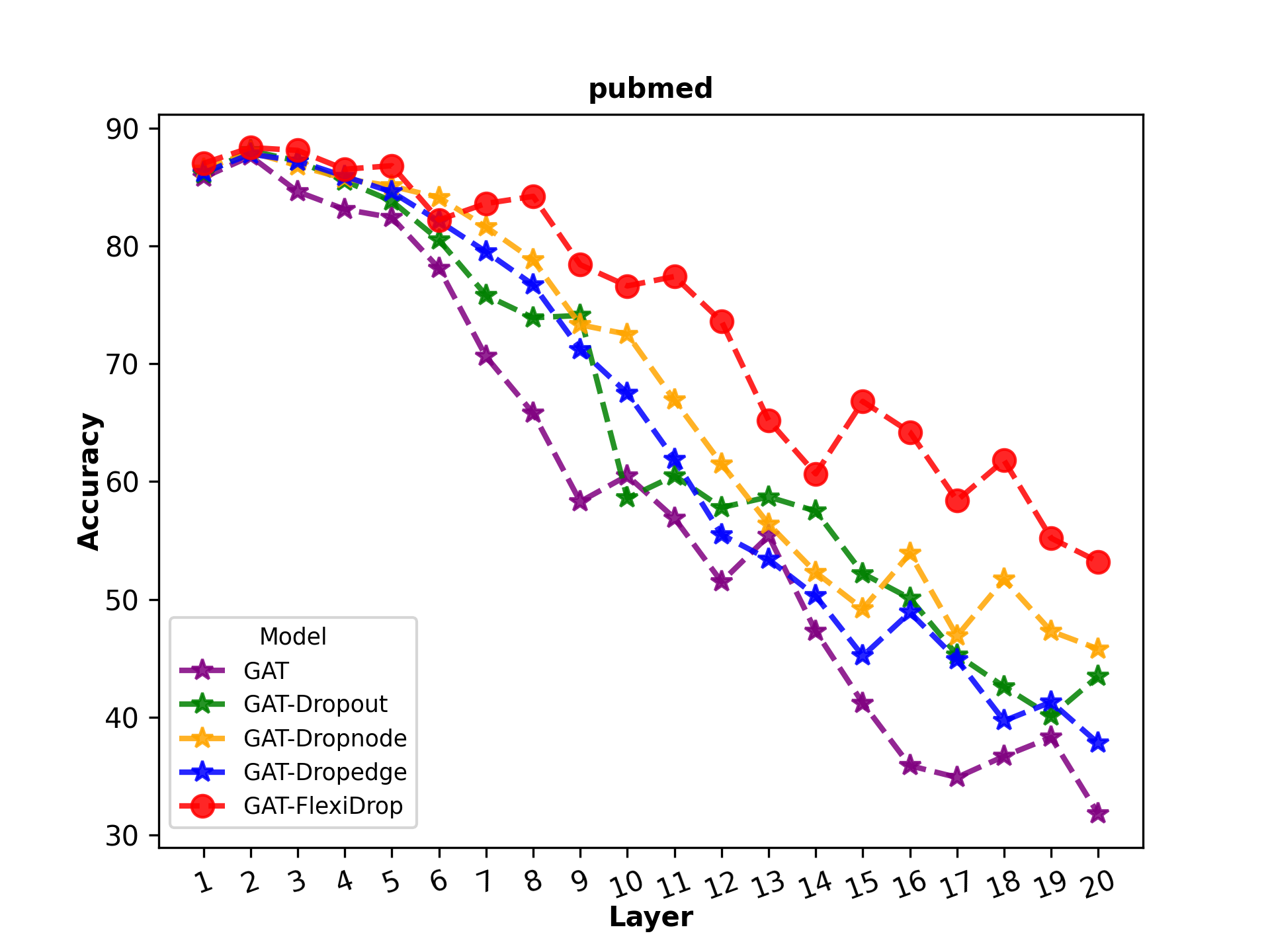}
	}
	\subfigure[$Dirichlet$ energy] {
		\centering
		\includegraphics[width=0.85\columnwidth]{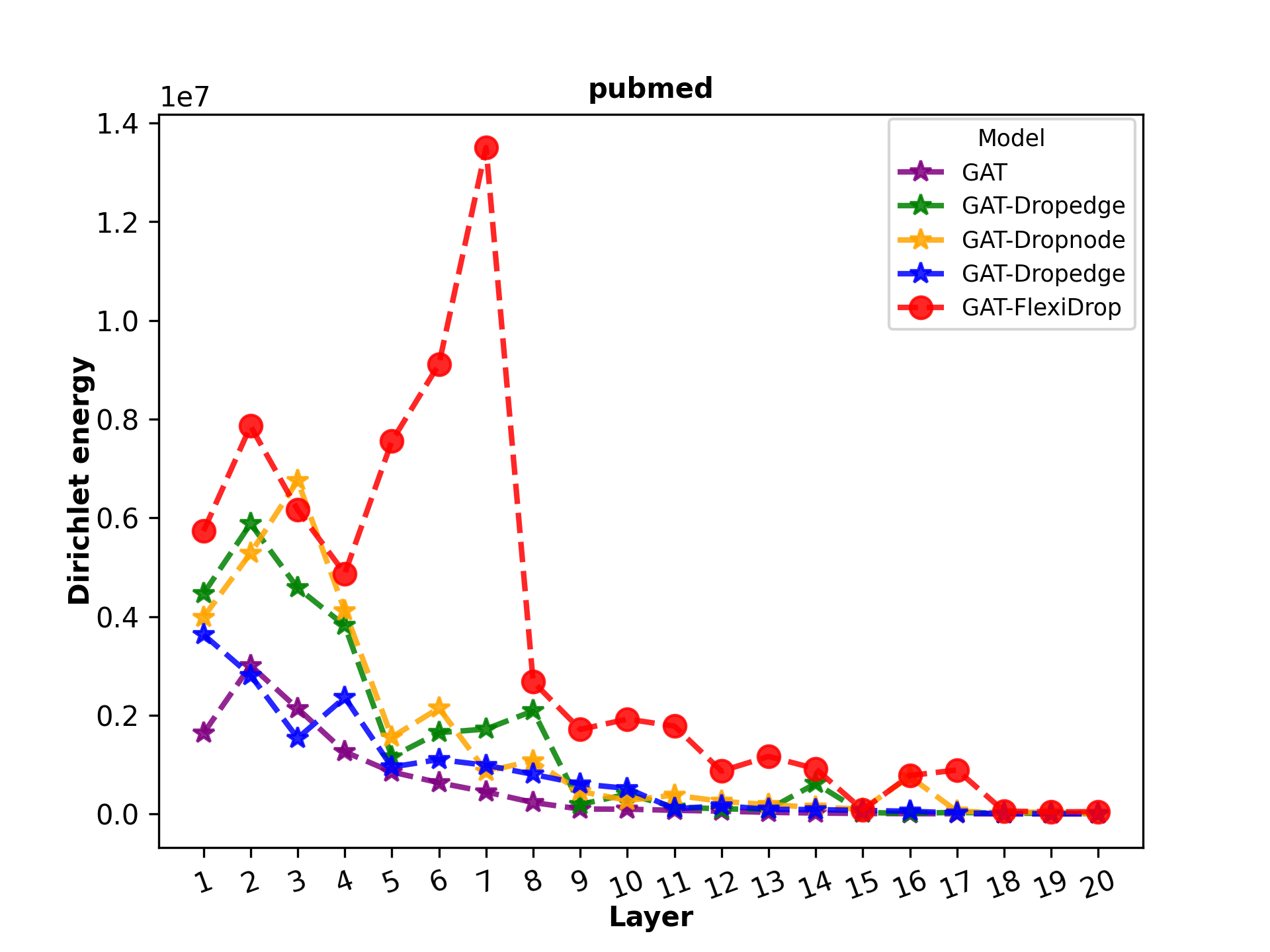}
  }

	\caption{The experimental results of over-smoothing analysis under node classification (where the task uses the PubMed dataset.)}\label{layer_analysis}
\end{figure*}

\paragraph{\textbf{Baseline methods and Backbone models}} In this paper, we compare our proposed FlexiDrop with Dropout~\cite{hinton2012improving}, Dropnode~\cite{feng2020graph}, and Dropedge~\cite{rong2019dropedge} methods. We adopt the dropout methods on various GNNs as the backbone model, and compare their performances on different datasets. We mainly consider three
mainstream GNNs as our backbone models: GCN~\cite{kipf2016semi}, GAT~\cite{velickovic2017graph}, and GraphSAGE~\cite{hamilton2017inductive}. 

\paragraph{\textbf{Parameter Configuration}}
For fair comparison, we stack two layers for all models. The hidden node embedding dimension of all the models is set to $256$ and the output embedding dimension is set to $32$. For the model optimization, all models will be trained $256$ epochs with a learning rate of $0.01$. For node classification task, the batch size is $1024$. For link prediction task, the batch size is $1024$. For graph classification task, the batch size is $32$. In performance experiments, we presented the results for other dropout methods using their optimal dropout rates. For FlexiDrop, the regularization coefficient $\lambda$ is set to 0.5. All models use Adam optimizer. In the experiments, all models are conducted with PyG and DGL platforms and PyTorch framework. See code at \href{https://github.com/Amihua/Flexi-Drop}{https://github.com/Amihua/Flexi-Drop}.

\subsection{Comparison Results}
In this section, we empirically validate the effectiveness of our proposed FlexiDrop. Table \ref{table_graph} and \ref{table2} summarize the overall results, with performance measured by accuracy. For the graph classification task, we use three public datasets (BZR, COX2, ENZYMES). And for node classification and link prediction tasks, we evaluate performance on three public datasets (Cora, CiteSeer, PubMed).  All results are averaged over 5 runs, with both the mean accuracy and standard deviation values reported.

\paragraph{\textbf{Performance in graph classification}} For graph classification, as shown in Table \ref{table_graph}, FlexiDrop excels on the BZR dataset with GAT, improving performance by $1.85 \%$ over Dropnode. Similarly, on the ENZYMES dataset with GraphSAGE, it outperforms Dropout by $4.19 \%$. 

\begin{table*}[h]
\centering
\normalsize
\caption{The experimental results of robustness analysis under node classification on citeseer dataset. The percentages in the table's header, from $10\%$ to $90\%$, indicate the number of edges randomly added to the dataset graphs. The best results are in bold.}
{
\begin{tabular*}{13.8cm}{c|ccccc}  
\hline  
 model & GCN & GCN-Dropout & GCN-Dropnode & GCN-Dropedge & GCN-FlexiDrop  \\  
\hline
   10\% & 77.79$\pm$0.38 & 78.25$\pm$0.53 & 78.43$\pm$0.58 & 78.03$\pm$0.32 & \textbf{78.65$\pm$0.27}  \\ 
 20\% & 75.41$\pm$0.33 & 76.56$\pm$0.61 & 76.65$\pm$0.49 & 76.25$\pm$0.75 & \textbf{77.76$\pm$0.43}  \\ 
 30\% & 73.91$\pm$0.43 & 74.26$\pm$0.54 & 74.76$\pm$0.77 & 73.62$\pm$0.62& \textbf{75.70$\pm$0.54} \\ 
   40\% & 72.03$\pm$0.24 & 72.89$\pm$0.49 & 73.70$\pm$0.83 & 72.27$\pm$0.57 & \textbf{73.93$\pm$0.63}  \\ 
50\% & 71.05$\pm$0.26 & 71.72$\pm$0.69 & 71.93$\pm$0.63 & 71.33$\pm$0.71 & \textbf{72.86$\pm$0.36}  \\ 
60\% & 68.96$\pm$0.48 & 70.17$\pm$0.31 & 70.48$\pm$0.51  & 69.13$\pm$0.80 & \textbf{71.54$\pm$0.21} \\  
   70\% & 67.28$\pm$0.46 & 68.76$\pm$0.77 & 68.98$\pm$0.75 & 67.89$\pm$0.53 & \textbf{69.70$\pm$0.35} \\ 
80\%  & 65.23$\pm$0.43 & 66.46$\pm$0.55 & 67.70$\pm$0.40 & 65.86$\pm$1.01 & \textbf{68.62$\pm$0.49}  \\ 
  90\%  & 63.87$\pm$0.34 & 65.32$\pm$0.87 & 65.74$\pm$0.62 & 64.72$\pm$0.81 & \textbf{66.74$\pm$0.52}  \\
\hline 
\end{tabular*}
}
\label{tab:robust}
\end{table*}

\begin{figure}[htbp]
	\centering
	\subfigure[Node Classification] {
		\centering
		\includegraphics[width=0.77\columnwidth]{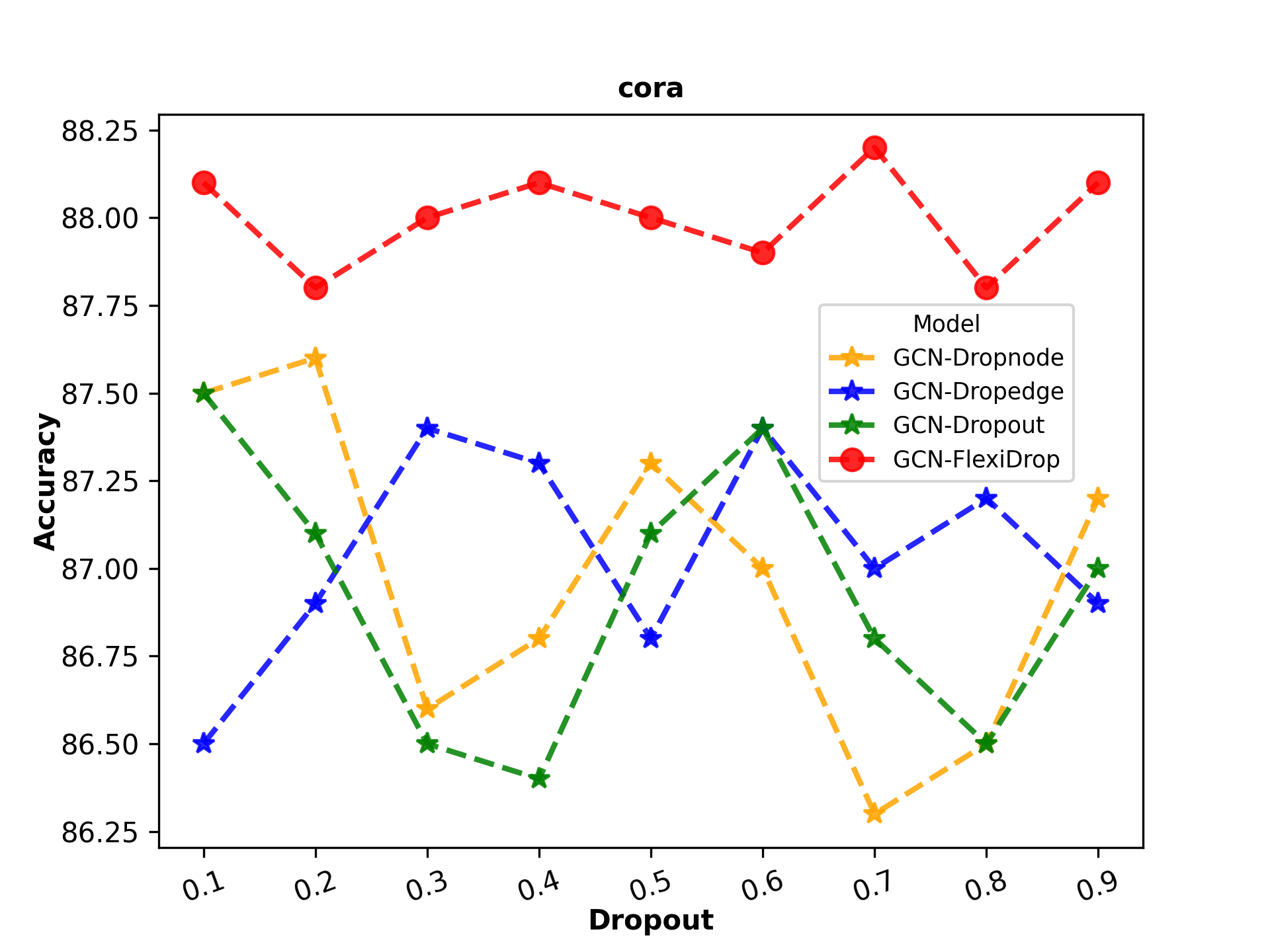}
	}
	\subfigure[Link Prediction] {
		\centering
		\includegraphics[width=0.77\columnwidth]{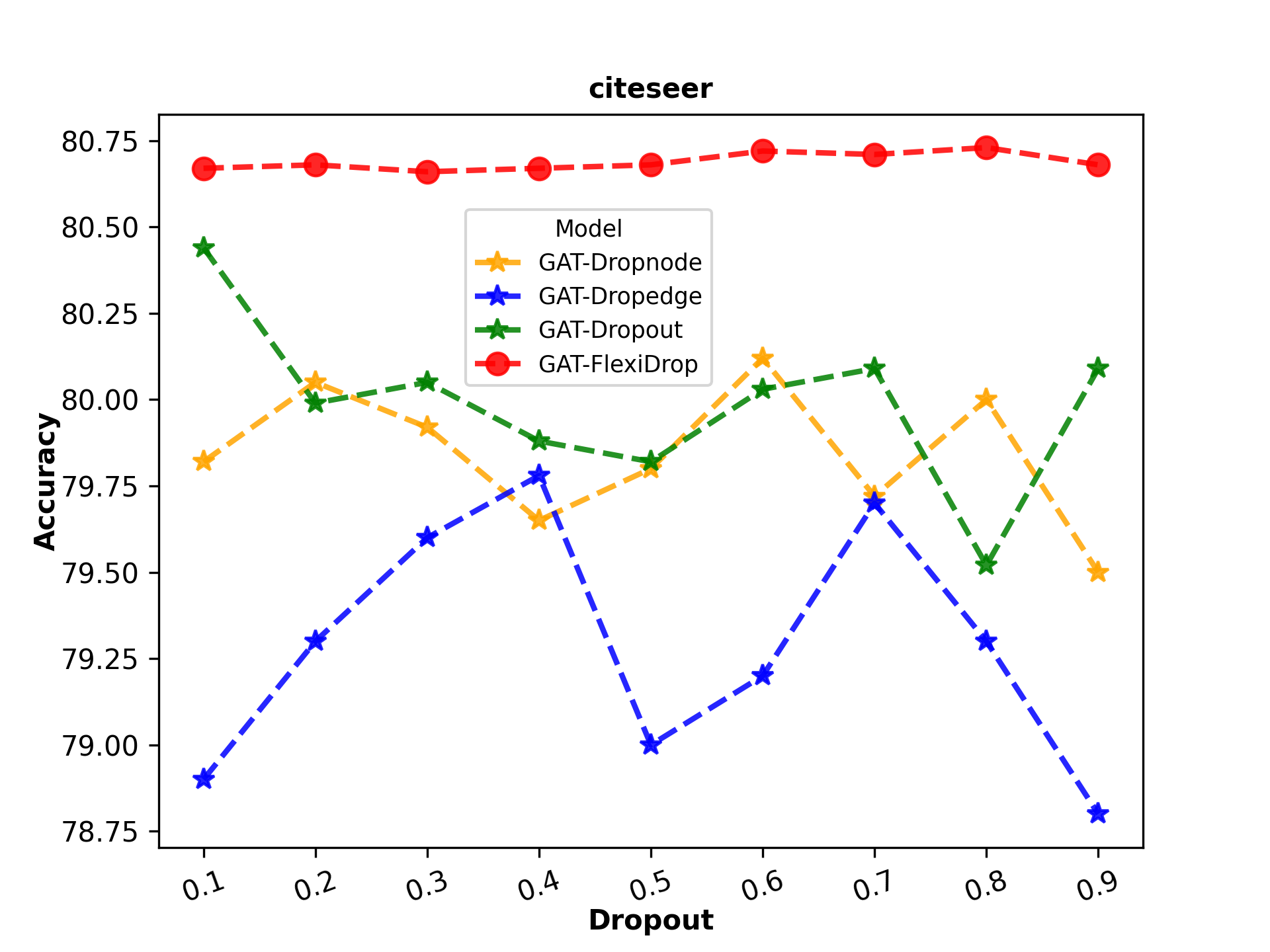}
	}
 \subfigure[Graph Classification] {
		\centering
		\includegraphics[width=0.77\columnwidth]{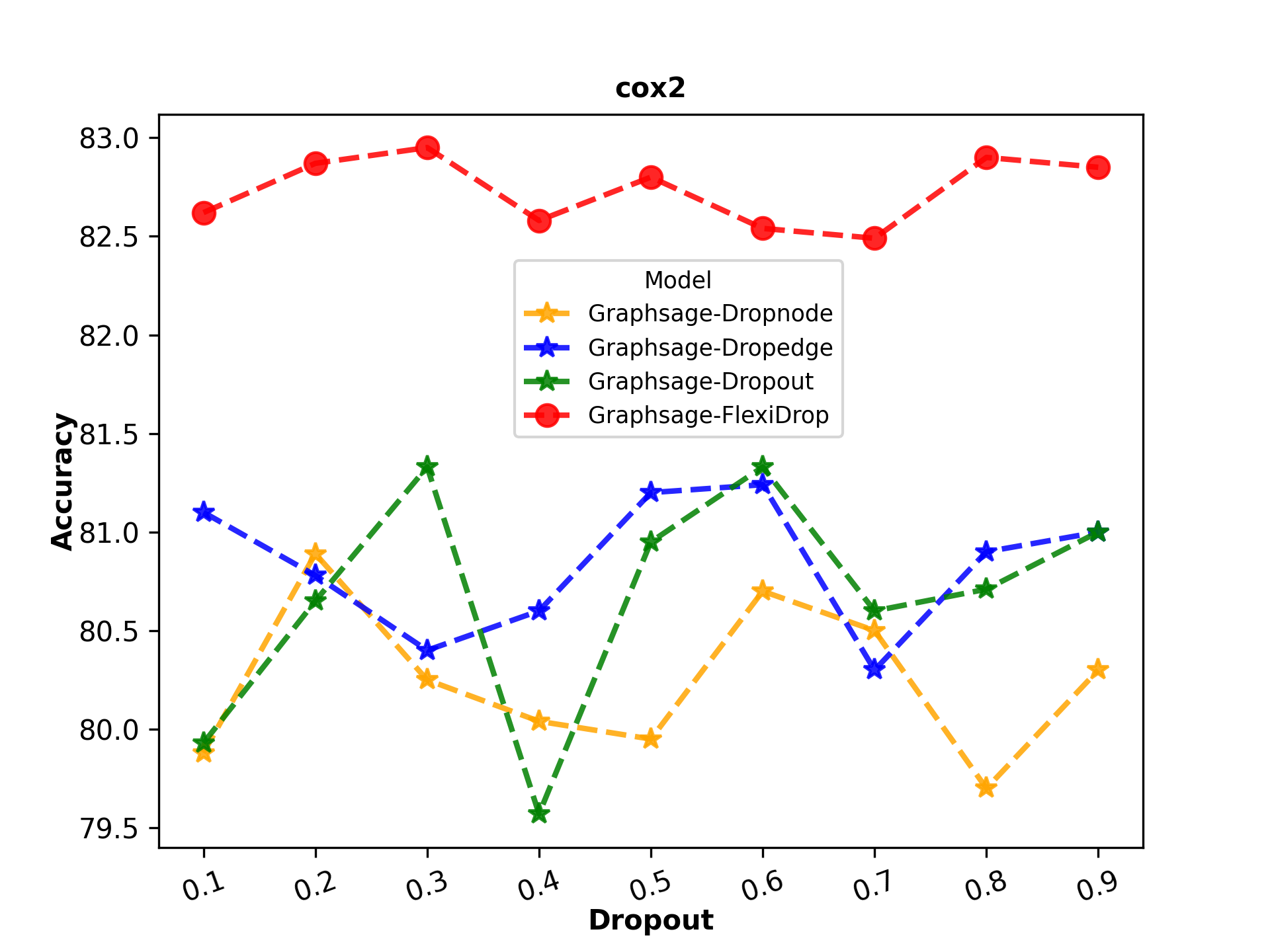}
	}
	\caption{The experimental results of parameter analysis under different tasks.
}\label{parameter_analysis}

\end{figure}

\paragraph{\textbf{Performance in node classification and link prediction}} From the results in Table \ref{table2}, we observe that FlexiDrop consistently outperforms random dropout methods across different GNN models and tasks (node classification and link prediction) on the Cora, CiteSeer, and PubMed datasets. For instance, in node classification with GAT on the Cora dataset, FlexiDrop achieves an accuracy of $87.27 \pm 0.47$, surpassing the next best method, Dropnode, by $1.52 \%$. In link prediction with GAT on the CiteSeer dataset, FlexiDrop attains $80.68 \pm 0.13$, representing a $0.70 \%$ increase over Dropnode.

Our results show that FlexiDrop provides more reliable and efficient message aggregation, delivering up to $4.19 \%$ better performance compared to other dropout methods across different datasets and models.

\subsection{Additional Results}
\paragraph{\textbf{Over-smoothing analysis}}
Experiments in Figure \ref{layer_analysis} reveal that FlexiDrop significantly outperforms dropout methods in mitigating over-smoothing in graph neural networks.
Figure \ref{layer_analysis}(a) shows that with increasing model depth, FlexiDrop consistently surpasses almost all dropout methods in terms of accuracy. This indicates that FlexiDrop is more effective at maintaining performance as model depth increases.

Adopting the approach from~\cite{dirichlet}, we used $Dirichlet$ energy at the $l^{th}$ layer
$$\mathcal{E}\left(\mathbf{x}^{(l)}\right):=\frac{1}{|\mathcal{V}|} \sum_{u \in \mathcal{V}} \sum_{v \in \mathcal{N}_u}\left\|\mathbf{x}_u^{(l)}-\mathbf{x}_v^{(l)}\right\|_2^2$$
to measure the model's over-smoothing.
Lower Dirichlet energy levels indicate greater over-smoothing in the model.
Figure \ref{layer_analysis}(b) shows the $Dirichlet$ energy across different layers for the same models and dataset. FlexiDrop maintains higher Dirichlet energy levels than the other methods, particularly as the number of layers increases. This suggests that FlexiDrop is more effective at preventing over-smoothing, and better preservation of feature distinctions across layers.

\paragraph{\textbf{Robustness analysis}} To validate the robustness of FlexiDrop, we conducted various levels of attacks on the edges of graphs in the node classification task. All the results are averaged across 5 repeat runs, std values of the experimental results are presented. We randomly added $10\%$ to $90\%$ more edges to the original graph and observed the changes in accuracy for the original GCN, GCN using dropout methods, and GCN employing FlexiDrop. The results presented in Table \ref{tab:robust} demonstrate that on CiteSeer dataset, GCN utilizing FlexiDrop outperforms the original models and those using dropout methods in terms of effectively resisting attacks, thereby exhibiting enhanced robustness. Remarkably, even under high noise attacks of $90\%$, our model maintains optimal accuracy performance, showcasing its ability to handle significant perturbations in graph structure without compromising effectiveness.

\paragraph{\textbf{Parameters analysis}}
We conducted an in-depth study on parameter sensitivity within dropout methods and FlexiDrop. We operate under the assumption that a model's performance, especially its accuracy, varies based on its sensitivity to parameter changes.
For each task type, we selected a benchmark dataset to evaluate the performance of various models employing either dropout methods or FlexiDrop. In dropout methods, we adjust the dropout rate $p$. Conversely, in FlexiDrop, we tweak the weight $\lambda$ of the loss function for parameter learning, since its dropout rate is adaptive. 

Figure \ref{parameter_analysis} shows that FlexiDrop achieves higher accuracy and greater stability across tasks. In Figure \ref{parameter_analysis}(a) for node classification on the Cora dataset, FlexiDrop consistently outperforms other methods across dropout rates. Figure \ref{parameter_analysis}(b) shows steady accuracy for link prediction on the CiteSeer dataset with FlexiDrop, compared to fluctuating results from other methods. For graph classification on the COX2 dataset in Figure \ref{parameter_analysis}(c), FlexiDrop also shows superior, more stable performance. 

These observations indicate that FlexiDrop, with its lower parameter sensitivity, is capable of adaptively learning the optimal dropout rate and dynamically adjusting to both the model and data. This adaptability leads to more robust and reliable performance across different tasks and dropout rates, showcasing the effectiveness of FlexiDrop in various scenarios.

\section{Conclusion}
In this paper, we propose a novel random dropout method for GNNs to achieve the adaptive adjustment of dropout rates. Based on rademacher complexity, we prove that the generalization error is bounded by a constraint function related to dropout rates for GNNs. By a theoretical derivation, we combine the constraint function and the empirical loss function and optimizing them simultaneously. As a result, we propose an adaptive random dropout algorithm for GNNs called FlexiDrop. By conducting experiments for multiple tasks on six public datasets, we demonstrate the effectiveness and generalization of our proposed method.

\section*{Acknowledgments}
This work is supported by National Natural Science Foundation of China (No.12326611) and the Fundamental Research Funds for the Central Universities, Nankai University (No.054-63241437).

% {\appendix[Proof of the Zonklar Equations]
% Use $\backslash${\tt{appendix}} if you have a single appendix:
% Do not use $\backslash${\tt{section}} anymore after $\backslash${\tt{appendix}}, only $\backslash${\tt{section*}}.
% If you have multiple appendixes use $\backslash${\tt{appendices}} then use $\backslash${\tt{section}} to start each appendix.
% You must declare a $\backslash${\tt{section}} before using any $\backslash${\tt{subsection}} or using $\backslash${\tt{label}} ($\backslash${\tt{appendices}} by itself
%  starts a section numbered zero.)}

%{\appendices
%\section*{Proof of the First Zonklar Equation}
%Appendix one text goes here.
% You can choose not to have a title for an appendix if you want by leaving the argument blank
%\section*{Proof of the Second Zonklar Equation}
%Appendix two text goes here.}

\bibliographystyle{IEEEtran}
\bibliography{ref}
\vfill
\end{document}